\documentclass[journal]{IEEEtran}
\usepackage{graphicx}
\usepackage{amsmath}
\usepackage{multirow}
\usepackage{color}
\usepackage{balance}
\usepackage{caption}
\usepackage{array}
\usepackage{threeparttable}
\usepackage{booktabs}
\usepackage{amsfonts}
\usepackage{bm}

\newcommand{\lwt}[1]{{\color{black} #1}}

\begin{document}
	\captionsetup{font={small}}
	
	\title{MARF: Multi-scale Adaptive-switch Random Forest \\
		for Leg Detection with 2D Laser Scanners}

	\author{Tianxi Wang,
		Feng Xue,
		Yu Zhou,~\IEEEmembership{Member,~IEEE,}
		and~Anlong Ming
		\thanks{T. Wang and F. Xue make equal contribution.}
		\thanks{This work was supported in part by the Major Project for New Generation of AI under Grant 2018AAA0100400, in part by the National Natural Science Foundation of China under Grant 62176098 and 61703049, in part by the Natural Science Foundation of Hubei Province of China under Grant 2019CFA022, and in part by the BUPT Excellent Ph.D. Students Foundation CX2020114. (Corresponding author: Yu Zhou)}
		\thanks{T. Wang, F. Xue and A. Ming are with \textcolor{black}{School of Computer Science}, Beijing University of Posts and Telecommunications, Beijing 100876, China
			(e-mail:\{wtx,xuefeng,mal\}@bupt.edu.cn).}
		\thanks{Y. Zhou is with School of Electronic Information and Communications, Huazhong University of Science and Technology, Wuhan 430074, China (e-mail: yuzhou@hust.edu.cn)}}
	
	\maketitle
	
	\begin{abstract}
		For the 2D laser-based tasks,
		e.g., people detection and people tracking,
		leg detection is usually the first step.
		Thus, it carries great weight in determining the performance of people detection and people tracking.
		However,
		many leg detectors ignore the inevitable noise and the multi-scale characteristics of the laser scan,
		which makes them sensitive to the unreliable features of point cloud and further degrades the performance of the leg detector.
		In this paper, 
		we propose a Multi-scale Adaptive-switch Random Forest (MARF) to overcome these two challenges.
		Firstly,
		the adaptive-switch decision tree is designed to use noise-sensitive features to conduct weighted classification and noise-invariant features to conduct binary classification,
		which makes our detector perform more robust to noise.
		Secondly,
		considering the multi-scale property that the sparsity of the 2D point cloud is proportional to the length of laser beams,
		we design a multi-scale random forest structure to detect legs at different distances.
		Moreover,
		the proposed approach allows us to discover a sparser human leg from point clouds than others.
		Consequently,
		our method shows an improved performance compared to other state-of-the-art leg detectors on the challenging Moving Legs dataset and retains the whole pipeline at a speed of \emph{60+} FPS on low-computational laptops.
		Moreover,
		we further apply the proposed MARF to the people detection and tracking system,
		achieving a considerable gain in all metrics.
	\end{abstract}

	\IEEEpeerreviewmaketitle

	\section{Introduction}
\begin{figure}[t]
		\includegraphics[width=0.97\linewidth]{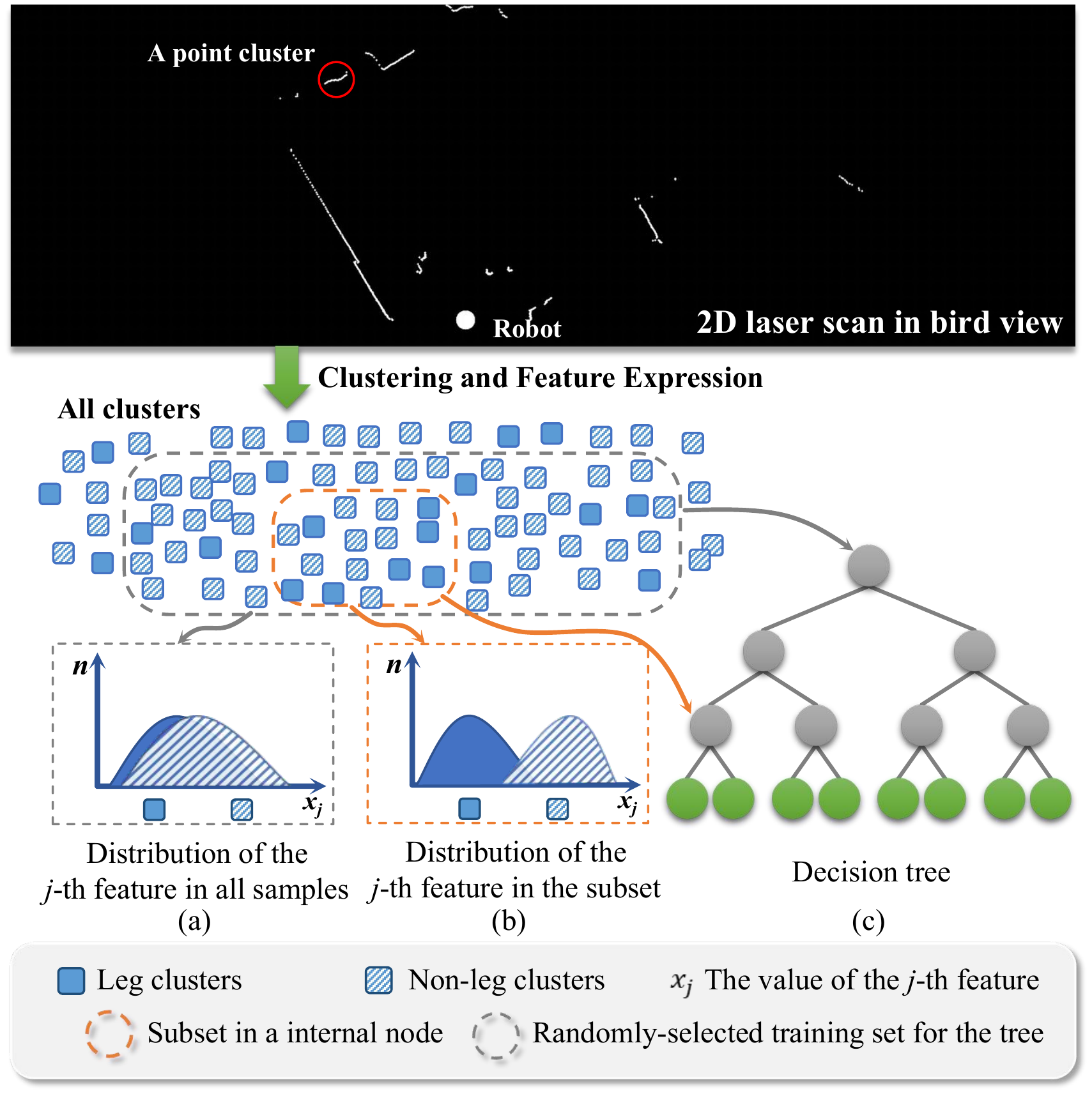}
		\caption{
			(a) depicts the value distributions of the $j$-th feature of all positive and negative samples.
			(b) depicts the value distributions of the $j$-th feature of a subset.
			(c) depicts a standard decision tree.}
		\label{fig:confidence}
	\end{figure}

	\IEEEPARstart{A}{s} the emergence of the various domain-specific
	requirements for the robot,
	such as security \cite{Lin2014Probability,FengXue-ICRA2019-TOD,FengXue-TIP2020-TOD},
	human-computer interaction (HCI) \cite{2007Ye,2018Yuan},
	visual following \cite{NIPS2012_3e313b9b,9605219,YuZhou-IJCV2016-SFVT},
	and navigation \cite{PAI20041025,2016Triebel},
	it is generally acknowledged that the people detection and tracking are necessary for robot \cite{2013Real,2013Gedik,2014ONLINE}.
	Recently,
	an increasing number of tasks \cite{Fod2002A,Beyer2018Deep,2003Zhao} utilize laser scanner to perceive people and the environment.
	However,
	as shown in Fig. \ref{fig:confidence},
	since the 2D laser scanner only measures the depth of a 2D plane above the ground,
	it is impossible to directly capture 3D human from 2D point clouds.
	To overcome this challenge,
	several previous methods mount the 2D laser scanner at the human leg's height,
	firstly detect each leg, and secondly match two legs to construct a human.
	Therefore,
	leg is considered as the input of these 2D laser-based people detection and tracking systems.
	For this reason,
	leg detector plays a dominant role in the performance of these systems.
	Next, we introduce the existing leg detector.

	Many leg detectors \cite{Chung2012The,2015Leigh,2018Li} employ a two-step leg detection approach as below:
	\begin{itemize}
		\item Firstly,
		a 2D laser scan is considered as a point cloud in 2D space,
		and is segmented into several point clusters by the distance jump between adjacent laser points \cite{2004Segment}.
		\item Secondly, several geometric features are employed to represent these atom clusters,
		so that a classifier is trained to classify each atom cluster into leg or non-leg.
	\end{itemize}
	Following this pipeline,
	Chung {\it et al}. \cite{Chung2012The} utilize the Support Vector Data Description to classify atom clusters.
	Li {\it et al}. \cite{2018Li} design several spatial relationship features and utilize AdaBoost \cite{Schapire1999A} to classify clusters.
	In recent years,
	random forest (RF) is widely employed to classify clusters due to its high efficiency and accuracy.
	It constructs many decision trees and adopts a bagging method to train each tree and aggregates the predictions of all trees.
	For clarity,
	RF is expressed as SRF below.
	Leigh {\it et al}. \cite{2015Leigh} and Linder {\it et al}. \cite{2016Linder} propose a SRF-based detector
	and apply it to build a robust people detection and tracking framework.
	However,
	the noise and the multi-scale property of laser scans are rarely discussed,
	which degrades the performance of these SRF-based leg detectors.
	We present them in detail below, respectively.

	{\bf Global-local confidence conflict:}
	We assume that the set of all clusters is expressed as $\mathbf{X}\in\mathbb{R}^{F\times N}$,
	consisting of the leg clusters $\mathbf{X}^+\in\mathbb{R}^{F\times N^+}$ and the non-leg clusters $\mathbf{X}^-\in\mathbb{R}^{F\times N^-}$,
	where $F$ is the dimensions of features and $N$ is the number of clusters, $N^+ + N^-  = N$.
	For the cluster set $\mathbf{X}$,
	we define the confidence of the $j$-th feature as the dissimilarity between the $j$-th feature distribution of the leg set $\mathbf{X}^+$ and that of the non-leg set $\mathbf{X}^-$,
	which is formulated as follows:
	\begin{equation}
		\mathbf{C}_j = 1 - \frac{min(\delta_{j}^+,\delta_{j}^-)}{max(\delta_{j}^+,\delta_{j}^-)},
		\left\{
		\begin{array}{rrr}
			\hspace{-1ex}\delta_{j}^+ = \frac{1}{N^+} \sum_i(\mathbf{X}^+_{i,j} - \overline{x_j^+})^2 \\
			\hspace{-1ex}\delta_{j}^- = \frac{1}{N^-} \sum_i(\mathbf{X}^-_{i,j} - \overline{x_j^+})^2
		\end{array}
		\right.
		\label{global confidance}
	\end{equation}
	where $i$ is the index of cluster and 
	$\overline{x_j^{+}}$ is the average of the $j$-th feature of leg clusters.
	Generally,
	a high confidence $\mathbf{C}_j$ indicates a high difference between the $j$-th feature of leg clusters and that of non-leg clusters,
	as shown in Fig. \ref{fig:confidence}(a).
	{During training a decision tree,
		the root node is trained to classify a sample set} $\mathbf{X}^*\subset\mathbf{X}$,
	and other internal nodes are trained to classify partial samples $\mathbf{A}\subset\mathbf{X}^*$.
	Assuming that the {\it global confidence} $\Phi_j$ and {\it local confidence} $\phi_j$ are the confidence of the $j$-th feature on set $\mathbf{X}$ and set $\mathbf{A}$,
	the nodes of tree always learn a feature has a maximum local confidence $\phi_j$,
	but ignore the low global confidence $\Phi_j$ caused by noise.
	If the selected feature has a low confidence $\Phi_j$,
	the global-local confidence conflict occurs:
	\begin{equation}
		\phi_j - \Phi_j \geq \epsilon , \epsilon \text{ is a threshold.}
		\label{eq:gl}
	\end{equation}
	In this case,
	the learned feature and the split point easily fail to classify samples out of the subset,
	and thus degrades the ability of the tree to recognize the leg cluster.
	As shown in Fig. \ref{fig:confidence},
	the $j$-th feature has strong discrimination on the positive and negative samples marked by orange circle well,
	but \lwt{an} extremely weak one on the samples marked by gray circle.
	This phenomenon is common due to the inevitable fluctuation of noisy laser data.

	{\bf Multi-scale characteristics:}
	Since the laser beam consists of many rays emitted from the center of the scanner,
	the farther the point cluster is, the sparser the cluster is,
	and vice versa.
	This scattering mechanism affects the value of some distance-sensitive features, such as the number of points,
	average inscribed angular,
	and distance to the scanner.
	However,
	existing methods usually classify multi-scale clusters by using a single-scale classifier,
	which degrades the leg detection performance in all scales.
	
	\textbf{Contributions:}
	Aiming to address these two issues and significantly boost the accuracy of the leg detector, 
	a novel Multi-scale Adaptive-switch Random Forest (MARF) is proposed in this paper.
	By evaluating the local and global confidence of the feature, an adaptive-switch decision tree is proposed.
	It employs an adaptive-switch strategy to decide whether a node of adapt is switched to a regular node or a dichotomous node,
	which alleviates the misclassification caused by the global-local confidence conflict.
	Moreover, 
	by taking the multi-scale characteristic of the leg features into consideration,
	a multi-scale structure is designed to detect the legs at multiple distances.
	Sufficient experimental results demonstrate the efficiency of the presented MARF.
	
	In summary, the key contributions of this paper lie in:
	
	1) The global-local confidence conflict of random forest is studied, which motivates us to introduce an adaptive-switch decision tree. It would conduct a weighted split if a conflict occurred, making the model more robust when using noise-sensitive features.
	
	2) A multi-scale random forest composed of adaptive-switch decision trees is introduced, which considers the multi-scale property of laser scan to improve leg detection accuracy in all scales.
	
	3) The proposed method outperforms all other leg detectors and \lwt{improves} the performance in the application of people detection and tracking. Meanwhile, it retains the whole leg detection pipeline at speed 60+ FPS on a low-computational laptop.

	\begin{figure*}[t]
		\centering
		\includegraphics[width=1\linewidth]{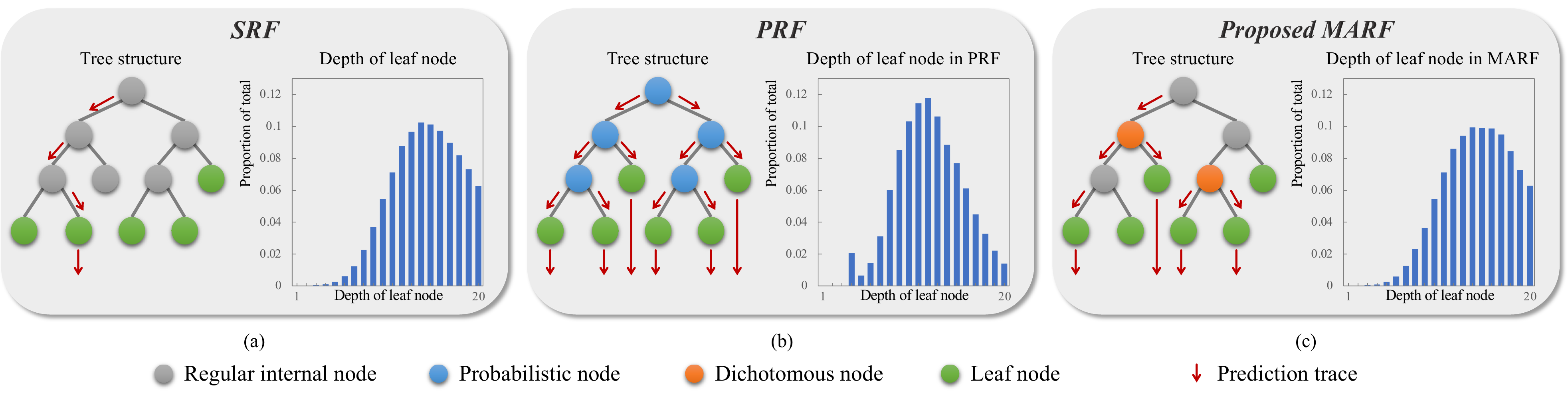}
		\caption{
			(a) depicts a standard decision tree and the depth of its leaf node.
			(b) corresponds to a probabilistic decision tree.
			(c) corresponds to our adaptive-switch decision tree.
			The red arrows in each tree indicate prediction trajectories of the cluster.
			The X-axis of the charts expresses the depth of the leaf node.
			The Y-axis of the charts expresses the proportion of leaf nodes of each depth to the total.}
		\label{fig:my_label}
	\end{figure*}
	
	\section{RELATED WORKS}
	\label{sec:related work}
	
	\subsection{Review of Leg Detection}
	As a fundamental task of people detection in 2D laser scanner,
	there have been several works on leg detection in past decades \cite{2008Spinello,2007Arras,2018Li,2020Gait}.
	These leg detectors can be classified into two types:
	simple heuristic methods and learning-based methods.
	
	Simple heuristic methods commonly use hand-crafted features and threshold conditions to distinguish leg and non-leg clusters.
	Moreover, the selection of thresholds requires experience.
	Xavier {\it et al}. \cite{2005FastLine} use legs-like parameters, e.g., diameter, to distinguish non-leg and leg.
	Similarly, Topp {\it et al}. \cite{Topp2005Tracking} propose to pick out some legs with blob-like \lwt{shapes},
	and Cui {\it et al}. \cite{2005Cui} pursue extracting moving leg blobs to detect and track stationary or moving people.
	These methods are simple but lack generalization in clutter scenarios.
	
	Learning-based methods couple hand-crafted features or deep features with machine learning approaches to build effective leg detectors and show excellent performance.
	Arras {\it et al}. \cite{2007Arras} utilize geometric features of clusters to train an AdaBoost \cite{Schapire1999A} classifier to detect legs.
	Weinrich {\it et al}. \cite{2014Weinrich} design distance-invariant \lwt{features} to describe clusters and train classifiers for people and wheelchairs detection.
	Based on prevailing works,
	Leigh {\it et al}. \cite{2015Leigh} combine features proposed in \cite{2012Arras} and \cite{Lu2013Towards} and employ SRF to classify legs and detect people.
	The above works aim at conducting the leg detection by different classifiers but ignore the inherent issues of LiDAR noise, occluded leg, etc., in the laser-based leg detection.
	To eliminate the misdetection of partially occluded legs,
	Li {\it et al}. \cite{2018Li} propose multi-type features to train strong AdaBoost classiﬁers.
	However, the complete lack of discussion in speed and noise impact restricts its practical value.
	Cha {\it et al}. \cite{cha_ijpem_2020} detect the human-leg in 3D feature space for a person-following mobile robot,
	which improves the recall of leg detection,
	but suffers from the precision drop due to lacking discriminative feature representation.
	Recently, Beyer {\it et al}. \cite{Beyer2017DROW} propose a novel deep learning method to detect people and wheelchairs and show remarkable performance.
	However, the deep learning-based methods can hardly strike a balance between efficiency and accuracy on the low-cost robot currently.
	Different from them,
	we focus on the rarely-discussed LiDAR feature noise,
	and propose a noise-immune random forest.
	Compared to our baseline \cite{2015Leigh},
	MARF improves the performance in all metrics.
	In addition, the proposed model can be directly trained based on noisy samples of other tasks, reflecting a better generalization of our model.


	\subsection{Review of Standard Random Forest}
	Standard random forest (SRF) is an ensemble learning method,
	which is utilized as a cluster classifier to divide point clusters into legs and non-legs in \cite{2015Leigh}.
	It aggregates the outputs of a set of standard decision trees \cite{Stone1984Classification} to improve the prediction accuracy \cite{2015Sen,2017Alejandro}.
	The structure of each decision tree is shown in Fig. \ref{fig:my_label}(a),
	which is a standard decision tree composed of regular internal nodes and leaf nodes.

	Each decision tree is trained by a randomly-selected subset from whole clusters training set $\mathbf{X}$.
	In detail,
	from the root of the tree,
	the regular internal node recursively splits the reached clusters by a {\it binary split function} below,
	and passes them to the left or right child node, formulating two subsets:
	\begin{equation}
		g(x_n,\tau_j) = [x_{n,j}<\tau_j]
		\label{eq:split function}
	\end{equation}
	where $x_{n,j}$ is the value of the $j$-th feature of the $n$-th cluster $x_n$,
	$\tau_j$ is a split point corresponding to the $j$-th feature,
	and $[\cdot]$ is the indicator function. 
	If $g(x_n,\tau_j) = 1$,
	cluster $x_n$ is passed to the left child node,
	otherwise right.
	
	To determine the optimal split point $\tau_j$ that separates leg and non-leg clusters as completely as possible,
	{\it Gini impurity} is employed to evaluate the impurity of each split subset:
	\begin{equation}
		G =\sum\nolimits_c \lambda^c(1-\lambda^c)
		\label{gini}
	\end{equation}
	where $\lambda^c$ is the percentage of $c$-labeled clusters among the subset.
	Traversing each feature and each split point, the regular internal node selects the $j$-th feature and split point $\tau_j$ that minimize the sum of impurity $G$ of the two split sets on the node as the optimal parameter.
	
	In this way,
	SRF learns optimal split point $\tau_j$ for internal nodes that seem to distinguish the clusters clearly.
	However,
	since some features are sensitive to the noise of laser data,
	the split point learned from the partial training set tends to deviate \lwt{from} the whole training set, which is not optimal.
	During prediction,
	the node utilizes improper split \lwt{points} to split clusters to the left or right,
	which is prone to assign clusters with wrong labels mistakenly.
	
	\subsection{Review of Probabilistic Random Forest}
	Considering that features may be affected by noise,
	different from SRF,
	probabilistic random forest (PRF) \cite{2019AJ} does not split the samples,
	but estimates a left-split probability and a right-split probability for each sample.
	Then it passes each sample to the left child node with the left-split probability and the right with the right-split probability simultaneously.
	As shown in Fig. \ref{fig:my_label}(b),
	the probabilistic node is designed to calculate the probabilities.
	Assuming that $x_n$ denotes the $n$-th reached sample on the current probabilistic node,
	the {\it probabilistic split function} of this node can be formulated as follows:
	\begin{equation}
		h(x_n,\tau_j) = \{p_b\cdot p_l,p_b\cdot p_r\}
		\label{eq:probabilistic split function}
	\end{equation}
	where $\tau_j$ is a split point corresponding to the $j$-th feature.
	$p_b$ is the basic probability of $x_n$,
	which is the probability passed from the parent node 
	and initialized with 1 at the root node.
	$p_l$ and $p_r$ are the left-split probability and right-split probability of $x_n$ at the current node, respectively.
	PRF assumes that each feature of sample $x_n$ contains a Gaussian noise.
	Then they \lwt{define} that $p_l$ and $p_r$ represent the probabilities of $x_{n,j}<\tau_j$ and $x_{n,j}\geq\tau_j$, respectively,
	where $x_{n,j}$ is the $j$-th feature of sample $x_n$, considered as a random variable.
	In more details,
	probability $p_l$ can be figured by the cumulative distribution function of Gaussian distribution $p_l=P\left(x_{n,j}\le \tau_j\right)$, and $p_r = 1-p_l$.
	Then,
	$p_b\cdot p_l$ and $p_b\cdot p_r$ are passed to the left child node and the right child node together with sample $x_n$, respectively.
	Note that $\tau_j$ is also determined by minimizing the sum of impurity of two sample subsets.
	But different from SRF,
	PRF defines $\lambda^c$ in Eq.$($\ref{gini}$)$ as $\lambda^c = \sum p^{c}_b/\sum p_b$ to measure the impurity of each split set,
	where $p^c_b$ indicates the basic probability of $c$-labeled sample in the current node. 
	
	Notably,
	as cluster $x_n$ goes deeper,
	the passing probability of $x_n$ sharply decreases.
	Once the $p_b$ approaches 0, namely lower than an eligible threshold,
	The training is terminated,
	and this node is transformed to a leaf node.
	As shown in the chart in Fig. \ref{fig:my_label}(b),
	the problem above is prone to generate lots of shallow-layer leaf nodes, which degrades the discrimination of the entire model.
	
	\begin{table}[t]
		\centering
		\caption{Feature Representation}
		\setlength{\tabcolsep}{5mm}
		\begin{tabular}{cc}
			\toprule
			\multicolumn{2}{c}{\textbf{Basic Geometric Features}}\\
			\midrule
			Width &  Circularity\\
			Linearity & Boundary length\\
			Number of points & Radius of best-fit circle\\
			Mean angular difference &  Standard distance to gravity\\
			Average distance to median & \\
			\midrule
			\multicolumn{2}{c}{\textbf{Relative Distance Features}}\\
			\midrule
			Mean curvature &   Boundary regularity\\
			Distance to scanner & Standard inscribed angular\\
			Average inscribed angular & Distance to scanner per point\\
			Left occlusion & Right occlusion \\
			\bottomrule
		\end{tabular}
		\label{tab:features}
	\end{table}
	
	\section{Problem Formulation}
	\label{sec:problem formulation}
	Following the prior methods \cite{2015Leigh,2007Arras},
	this paper decouples leg detection into two steps:  
	points clustering and leg classification.
	In the first step,
	given a 2D laser scan as input,
	the laser scan is clustered into a set of point clusters according to the jump distance between adjacent laser points \cite{2015Leigh},
	and each point cluster is a candidate for leg classification.
	In the second step,
	each cluster is expressed by utilizing several features \cite{2012Arras,Lu2013Towards},
	as shown in Table \ref{tab:features}.
	And the cluster is classified into two types, namely, leg or non-leg.
	Typically,
	there are two types of features: {\it basic geometric features} and {\it relative distance features}.
	Basic geometric features describe the geometric shape parameters of the cluster.
	Relative distance features describe the distance between adjacent points of the cluster, are more sensitive to the distance variance.
	In detail,
	let $x_n\in\mathbb{R}^{17}$ denotes the feature vector of the $n$-th cluster,
	all feature vectors are stacked as the whole training set $\mathbf{X}\in\mathbb{R}^{N\times17}$.
	The corresponding labels are denoted as $\mathbf{Y}\in\mathbb{R}^N$,
	i.e., $+1$ for leg clusters and $0$ for non-leg clusters.
	
	\section{METHOD}
	\label{sec:method}
	In this section,
	our method is described in two levels:
	tree level and forest level.
	We first propose the adaptive-switch decision tree to overcome the global-local confidence conflict (see Sec. \ref{sec:adaptive-switch}).
	Secondly, the multi-scale adaptive-switch random forest (MARF) is proposed to fully utilize the multi-scale characteristics of \lwt{the} human leg in the 2D space (see Sec. \ref{sec:multi-scale}).

	\subsection{Adaptive-switch Decision Tree}
	\label{sec:adaptive-switch}
	
	To address the global-local confidence conflict,
	we propose the adaptive-switch decision tree.
	It consists of three types of \lwt{nodes}:
	1) regular internal nodes,
	2) dichotomous nodes,
	and 3) leaf nodes.
	Each regular node receives clusters from the parent node and splits them into two subsets by the noise-insensitive feature,
	and passes them to the left child node and the right child node.
	Unlike the regular node,
	each dichotomous node calculates two weights for each cluster by the noise-sensitive feature,
	and then passes all clusters to both child nodes.
	Then, we give the description of training this model below.
	
\begin{figure}[t]
		\centering
		\includegraphics[width=1\linewidth]{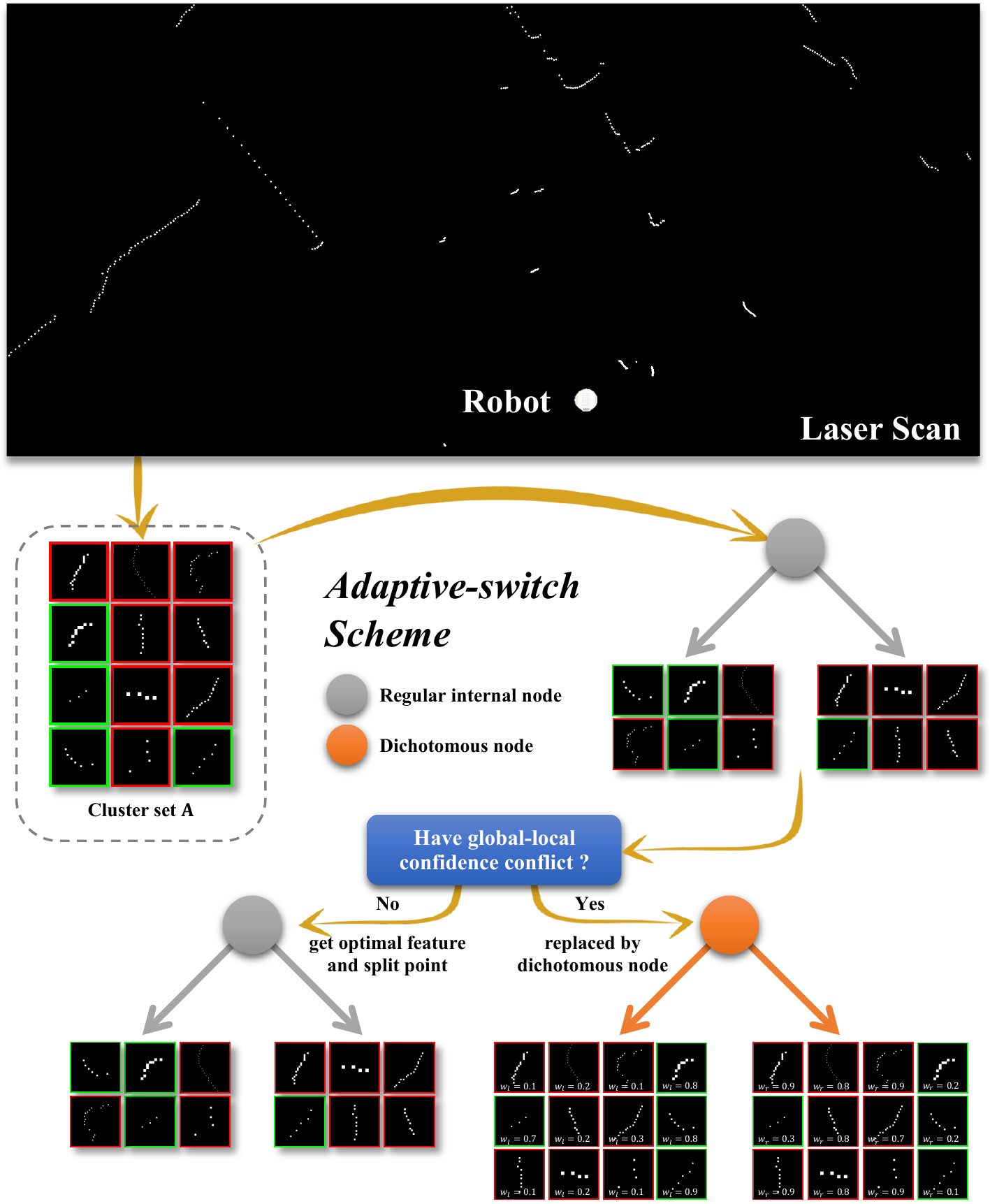}
		\caption{The pipeline of training a single node. The clusters marked in red are non-leg clusters, and green for leg clusters.}
		\label{fig:meth1}
	\end{figure}

	Specifically,
	given a randomly-selected set $\mathbf{X_t}\subset \mathbf{X}$ for training the tree,
	starting from the root node,
	each node is trained as follows.
	We assume that $\mathbf{A_t}\subset \mathbf{X_t}, \mathbf{A_t}\in \mathbb{R}^{N_\mathbf{A}\times17}$ is the cluster set reaching on a specific node.
	At the beginning of training,
	this node is initially treated as a regular internal node.
	Randomly selecting $\lfloor\sqrt{17}\rfloor$ features as the candidates, wherein $\lfloor\cdot\rfloor$ denotes the floor operation,  
	this node obtains the feature and the split point with minimum Gini impurity (Eq.$($\ref{gini}$)$) to split the clusters $\mathbf{A_t}$.
	Then,
	as shown in Fig. \ref{fig:meth1},
	if the confidences of the selected $j$-th feature in $\mathbf{X_t}$ and $\mathbf{A_t}$ satisfy the conflict as Eq.$($\ref{eq:gl}$)$,
	namely, the global-local confidence conflict occurs,
	a dichotomous node is employed to replace this regular internal node,
	and the optimal parameters are re-selected to alleviate the conflict.
	This type of node is described below.

	\begin{figure*}[t]
		\centering
		\includegraphics[width=1\linewidth]{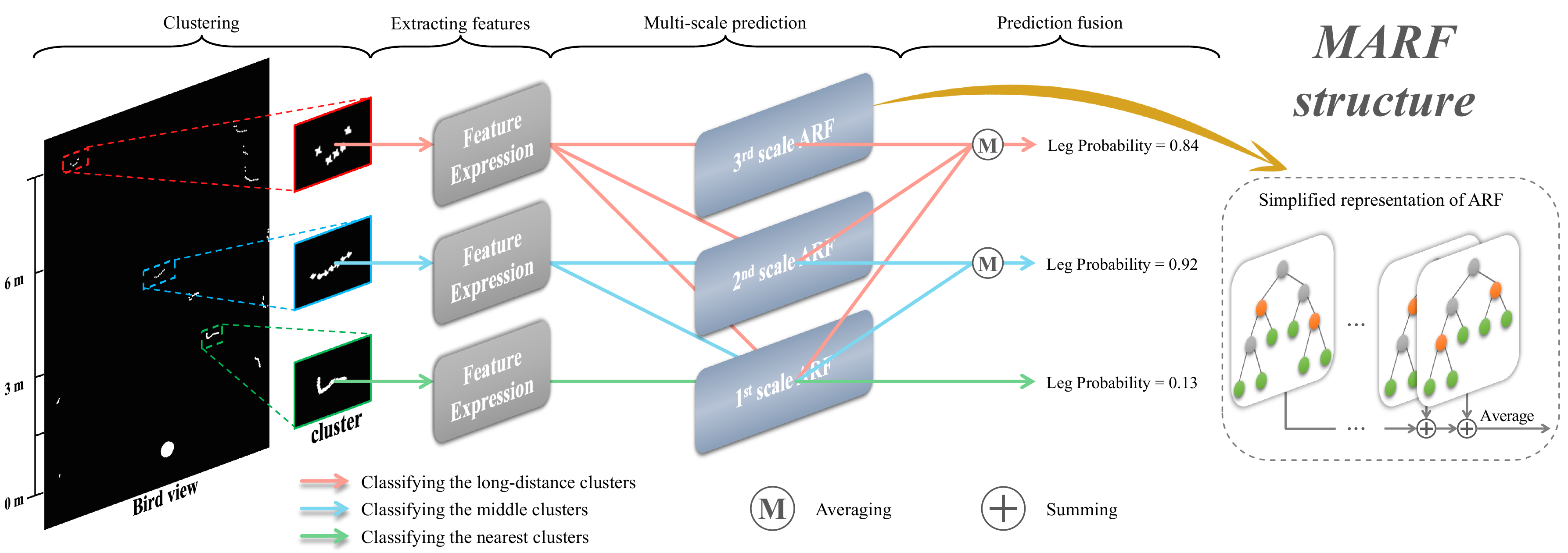}
		\caption{A simplified model shows the prediction of MARF.
			Arrows with different colors indicate different forest combinations for the corresponding color clusters.
			The near clusters are classified by forest $F^{(1)}$,
			and the clusters at a middle distance are classified by forest $F^{(1)}$ and $F^{(2)}$.
			Analogically, the far clusters correspond to forest $F^{(1)}$, $F^{(2)}$, and $F^{(3)}$.
			The red dotted box shows the combination rule of the operator $\oplus$.}
		\label{fig:multi_layer}
	\end{figure*}

	\textbf{Dichotomous Node:} We propose the dichotomous node to alleviate the global-local confidence conflict in this work.
	Different from the regular internal node,
	the dichotomous node calculates two weights of each cluster $x_n\subset\mathbf{A_t}$
	by the {\it dichotomous split function} as follows:
	\begin{equation}
		d(x_n,\tau_j) = \{w_l,w_r\}
		\label{dichotomous split function}
	\end{equation}
	where $j$ indicates the $j$-th feature selected by the regular internal node before the node is \lwt{converted} to the dichotomous node. $\tau_j$ is the re-selected split point corresponding to the $j$-th feature.
	Noteworthy,
	due to the effect of noise on features,
	we make a hypothesis that the $j$-th feature $x_{n,j}$ 
	has a Gaussian noise with variance $\sigma_j^2$,
	i.e., $x_{n,j} \sim \mathcal{N}(x_{n,j},\sigma_j^2)$,
	where $\sigma_j^2$ is the variance of the $j$-th feature in set $\mathbf{X}$.
	More experimental discussion about the Gaussian noise hypothesis can be found in the supplementary material.
	Based on this hypothesis,
	$w_l$ is defined as the cumulative distribution of $x_{n,j}$ at split point $\tau_j$.
	And $w_r$ is the complementary cumulative distribution of $x_{n,j}$ at point $\tau_j$, i.e., $w_l + w_r =1$.
	They are regarded as the weights of cluster $x_n$ passed to the left and right child nodes,
	respectively,
	and they are formulated as follows:
	\begin{flalign}
		\begin{split}
			w_l = \frac{1}{\sigma_j\sqrt{2\pi}}\int_{-\infty}^{\tau_j}\textrm{exp}\Big ( -\frac{(z-x_{n,j})^2}{2\sigma_j^2}\Big )dz\\
			w_r = \frac{1}{\sigma_j\sqrt{2\pi}}\int_{\tau_j}^{+\infty}\textrm{exp}\Big ( -\frac{(z-x_{n,j})^2}{2\sigma_j^2}\Big )dz
		\end{split}
		\label{pl}
	\end{flalign}
	where $z$ denotes the variable of integration.
	
	To learn the split point $\tau_j$ that achieves the minimum summed impurity of the two split subsets, consisting of weighted clusters,
	we calculate the \textit{Gini impurity} as follows.
	Assuming that the leg cluster corresponds to the superscript `$+$' and \lwt{the} non-leg cluster corresponds to `$-$',
	the \textit{Gini impurity} of left or right sets is calculated as follows:
	\begin{equation}
		G_{l/r} = 2\times\frac{\sum\nolimits w_{l/r}^{+}}{\sum w_{l/r}}\times\frac{\sum\nolimits w_{l/r}^{-}}{\sum w_{l/r}}
	\end{equation}
	where $w^+$ denotes the weights of the leg clusters,
	$l$ and $r$ indicate the left set and the right set.
	Furthermore,
	all split points of the $j$-th feature are tried to minimize the summed impurity of \lwt{the} left set and right set:
	\begin{equation}
		\bm{\tau}_j = \mathop{\bm{{\arg\min}}}_{\tau_j}\big(G_l\times\sum\nolimits w_{l} +  G_r\times\sum\nolimits w_{r}\big)
		\label{eq:cg}
	\end{equation}
	
	Finally,
	the dichotomous node stores the selected $j$-th feature and the optimal split point $\bm{\tau}_j$,
	then passes all clusters to both left child nodes.
	Furthermore,
	the left weight $w_l$ are passed to the left child node,
	and $w_r$ corresponds to the right child node,
	as shown in Fig. \ref{fig:meth1}.

	In this way,
	each regular internal node and dichotomous node and trained to construct an adaptive-switch decision tree.
	In addition,
	the node stops training and turns into a leaf node if any of the following constraints are met:
	\begin{itemize}
		\item The cluster number of set $\mathbf{A_t}$ satisfies $N_\mathbf{A} < 2$.
		\item The depth of the node is higher than $20$.
		\item The summed Gini impurity of the node is less than $e^{-6}$
	\end{itemize}
	Each leaf node saves the dominant label of clusters reaching on this node.

	\textbf{Prediction of a single tree}:
	Assuming that $x_i$ denotes the feature vector of a point cluster,
	and $f$ denotes a trained adaptive-switch decision tree,
	we use the model $f$ to predict the label of the cluster $x_i$.
	The leaf nodes this cluster $x_i$ reached are denoted as $\{l_q\}, l_q\in\{0,1\}, q\in\{1,2,...,Q\}$,
	and the weights of $x_i$ partitioned by the dichotomous nodes on the path from the root to the $q$-th leaf nodes are denoted as $\{w_e^q\}, w_e^q\in[0,1], e\in\{1,2,...,E\}$,
	where $Q$ is the number of leaf nodes and $E$ is the number of dichotomous nodes that the sample $x_i$ passes from the root node to $q$-th leaf node.
	The prediction of the tree $f$ is formulated as:
	\begin{equation}
		f(x_i) = 
		\begin{cases}
			1, &\sum\nolimits_{q=1}^Q \big(\prod\nolimits_{e=1}^E(w_e^q)\times l_q \big)>0.5\\
			0, &\text{otherwise}
		\end{cases}
	\end{equation}
	where $f(x_i)$ denotes the label of cluster $x_i$ predicted by $f$.
	Compared to SRF,
	the proposed tree no longer decisively classifies clusters,
	but jointly considers the weights of the labels on multiple leaf nodes,
	avoiding wrong splitting caused by the noise-sensitive feature.

	\textbf{Differences from PRF:}
	Although our method is inspired by PRF \cite{2019AJ},
	there are considerable differences between them.
	Firstly,
	PRF conducts probabilistic classification on all nodes, 
	but our method only conducts weighted split when a noise-sensitive feature is inevitably selected,
	and the global-local confidence conflict occurs.
	Secondly,
	during training a node,
	PRF computes the probability of each sample by multiplying all the probabilities on the path from the root to this node,
	causing a sharp drop in the basic probabilities and premature termination of training.
	In contrast,
	our method re-balances the weights for the clusters \lwt{that} reached the current node before the split.
	The two differences make our methods fully trained,
	and improve the performance of \lwt{the} leg detector.
	
	\subsection{Multi-scale Adaptive-switch Random Forest}
	\label{sec:multi-scale}
	
	Recently, the multi-scale strategy has been proven to be very useful for detection in point clouds \cite{Qi2017PointNet} and image \cite{WANG2021107593}.
	To discover the long-distance legs,
	we design the multi-scale adaptive-switch random forest (MARF).
	It utilizes the adaptive-switch decision tree and the multi-scale property of the point cluster.
	In this section,
	we first introduce the discrete 2D space to determine the scale of each cluster,
	then describe the structure and the training scheme of MARF.
	
	\textbf{Spatial Discretization:}
	To determine the scale of each cluster,
	we first discrete the depth interval $[0m,+\infty)$ into $K$ sub-intervals,
	where $K \in \mathbb{N}^+$.
	Since SRF hardly distinguishes the clusters at a distance larger than $6m$,
	we assign the point clusters further than $6m$ into one scale.
	Then we partition range $[0m,6m)$ to determine other scales.
	In this way,
	the 2D space is discretized into several depth intervals.
	For a two-scale model,
	the space is discretized into $[0m,6m),[6m,+\infty)$.
	And $[0m,3m),[3m,6m),[6m,+\infty)$ for three-scale model,
	$[0m,1.5m),[1.5m,3m),[3m,6m),[6m,+\infty)$ for four-scale model.
	All clusters in $k$-th scale are denoted as $\{\mathbf{X}_k|\mathbf{X}_k\in\mathbb{R}^{N_k\times17}, k\in\{1,2,...,K\}\}$

	\textbf{Structure:}
	The MARF consists of several adaptive-switch decision trees.
	Corresponding to the number of scales $K$,
	there are $K$ sets of trees in MARF,
	namely, the adaptive-switch random forest (ARF),
	as shown in Fig. \ref{fig:multi_layer}.
	In detail,
	the $k$-th ARF contains $T_k$ trees.
	To recognize the long-distance leg as much as possible,
	the $k$-th ARF is utilized to classify the point clusters inside the $k$-th scale.
	For instance,
	when $K = 3$,
	the $1$-th ARF classifies clusters inside the range $[0m,+\infty)$,
	the $2$-th ARF classifies clusters inside the range $[3m,+\infty)$,
	and the $3$-th ARF corresponds to the range $[6m,+\infty)$.
	The training scheme is described below.
	
	\textbf{Multi-scale Training Scheme:}
	Let $\mathbf{F} = \{F^{(k)}\}, k\in\{1,2,...,K\}$ denotes the MARF with $K$ scales,
	where $F^{(k)}$ is the $k$-th ARF in MARF.
	To capture the long-distance leg that is hardly detected,
	we adopt a biased sampling strategy to construct the training set for each tree during the training.
	More specifically,
	
	we randomly select clusters from set $\mathbf{X}$ with replacement.
	For the randomly-selected cluster in the set $\mathbf{X}_k\cup\mathbf{X}_{k+1}\cup...\mathbf{X}_K$,
	it is accepted as a training sample.
	To avoid overfitting caused by the lack of scale variety,
	for the randomly-selected cluster in the set $\mathbf{X}_1\cup\mathbf{X}_{2}\cup...\mathbf{X}_{k-1}$,
	it is accepted with probability $\alpha$.
	In this way,
	we construct an independent training set that contained $N$ clusters for each decision tree in $F^{(k)}$.
	Then, each decision tree is constructed by following the way described in Sec. \ref{sec:adaptive-switch}.
	
	Since ARF $F^{(1)}$ learns the leg clusters of all scales,
	it can be used to predict \lwt{the} legs of all scales.
	In contrast,
	ARF $F^{(2)}$ focuses on learning the farther legs, further alleviating the false-positive detection of long-distance legs.
	And so on,
	ARF $F^{(K)}$ has a more robust capability to distinguish the clusters inside the $K$-th scale.
	Thus, this ARF is only employed to predict \lwt{the} farthest clusters.

	\textbf{Leg Prediction:}
	To recognize the legs in all scales,
	we introduce an overlapping fusion strategy to fuse all ARF predictions.
	Fig. \ref{fig:multi_layer} illustrates the prediction of a three-scale MARF.
	For a point cluster $x$ in the $k$-th scale,
	ARFs $\{F^{(\mathbf{k})}|\mathbf{k}\in\{1,...,k\}\}$ are employed to predict its probability of being a leg.
	Each ARF outputs the average value of the outputs of all decision trees within it.
	At last, if cluster $x_n$ is predicted by number $k$ scales ARF, MARF predicts the probability that the cluster belongs to the leg, can be formulated as follows:
	\begin{equation}
		\mathbf{F}(x_n) = \frac{1}{k} \sum_{\mathbf{k}=1}^{k}\big(F^{(\mathbf{k})}(x_n)\big)
	\end{equation}
	In our method,
	if $\mathbf{F}(x_n)$ is higher than $\beta$,
	the cluster $x$ is predicted as a leg cluster,
	where $\beta$ is a threshold for label assignment.
	
	\section{Experiments}
	\label{sec:experiments}
	
	In this section, the performance of multi-scale adaptive-switch random forest is evaluated and analyzed in detail.
	The experimental settings are firstly demonstrated in Sec. \ref{sec:dataset}.
	Next in Sec. \ref{sec:rf_method}, 
	the results of other leg detectors are compared with our method.
	Then, the ablation study is given in Sec. \ref{sec:ablation} to illustrate the significance of each design.
	Furthermore, in Sec. \ref{details}, we conduct analysis and comparative experiments to represent the effectiveness of details in our method.  
	Finally,
	we apply our method to a people detection and tracking framework to study its advancement in Sec. \ref{further}.

	\begin{figure}[t]
		\begin{center}
			\includegraphics[width=1\linewidth]{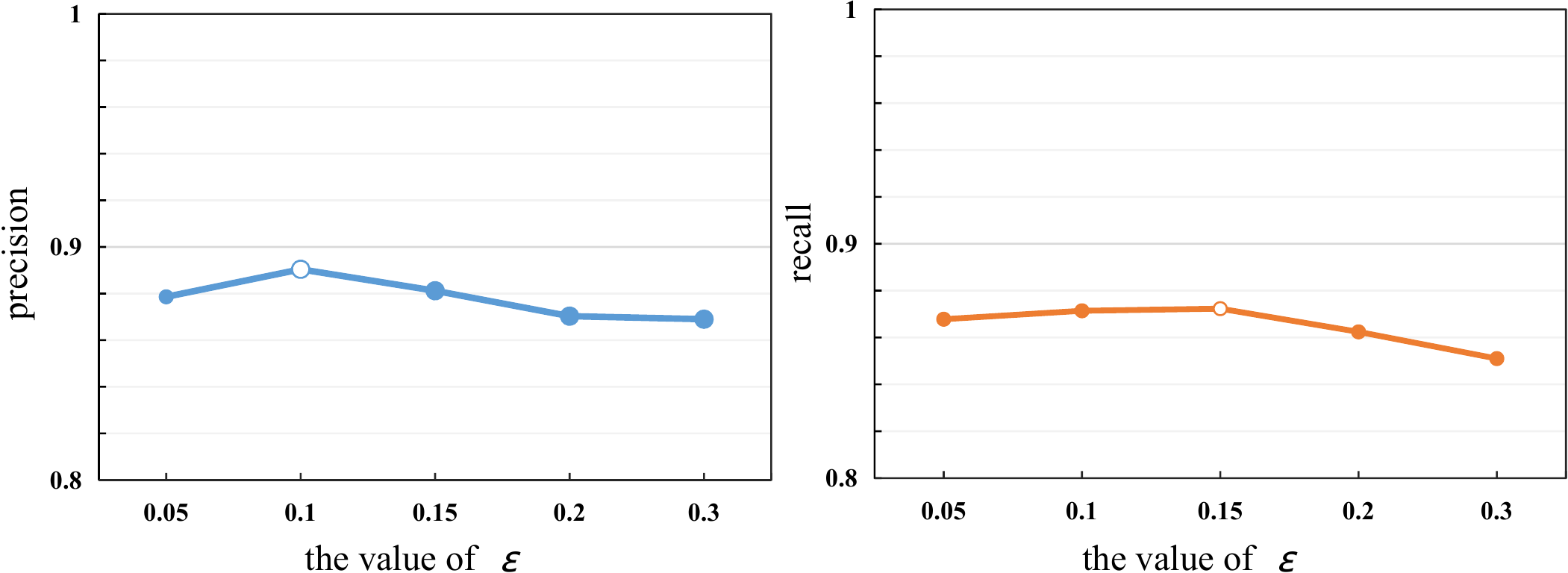}
			\caption{The precision of leg prediction  with different settings of parameter $\epsilon$.}
			\label{parameter}
		\end{center}
	\end{figure}
	
	\subsection{Experimental Settings}
	\label{sec:dataset}
	\subsubsection{Datasets}
	{\it Moving Legs} dataset is collected from the {\it People Tracking} dataset \cite{2015Leigh} to evaluate the performance of leg detection.
	The laser scan of the {\it Moving Legs} dataset is the same as the {\it People Tracking} dataset.
	But the label of {\it Moving Legs} dataset is extracted from {\it People Tracking}.
	In detail,
	the dataset labels the location of each people in the laser scan.
	Following the criterion utilized in \cite{2015Leigh},
	we define the true leg cluster as each two closest clusters near each ground truth people,
	with the condition that the distance between the leg cluster and the location of the people should be less than 0.5m.
	In this way,
	the collected dataset contains 26344 legs and 72251 non-legs.
	In addition,
	since part of scenarios are only labeled with the positions of people tracked,
	hence,
	only leg clusters are collected from those laser scans.
	Generally,
	clusters collected in this dataset are all in the real environment.
	Therefore,
	the evaluation results on this dataset can significantly reflect the performance of leg detection.
	
	{\it People Tracking} is a ROS-enable dataset provided by \cite{2015Leigh},
	which is the only public and available dataset for 2D laser-based people tracking as we know.
	This dataset is captured by a laser scanner with 0.35$^\circ$ angle resolution and 15m maximum detection distance.
	Moreover, this dataset is collected in the scenarios with single and multiple people tracked, and those scenarios are further employed to evaluate the contribution of our approach for people detection and tracking:
	\begin{itemize}
		\item SCENARIO \uppercase\expandafter{\romannumeral1}: {\it Multiple People, Stationary Robot}.
		\item SCENARIO \uppercase\expandafter{\romannumeral2}: {\it Multiple People, Moving Robot}.
		\item SCENARIO \uppercase\expandafter{\romannumeral3}: {\it Single People, Moving Robot}.
	\end{itemize}
	wherein long-distance people and crowds are the main challenges for people detection and tracking in SCENARIO \uppercase\expandafter{\romannumeral1}.
	In SCENARIO \uppercase\expandafter{\romannumeral2}, the robot motion induces more fluctuations, which further increases leg detection difficulty.
	In SCENARIO \uppercase\expandafter{\romannumeral3}, only the tracked person is labeled, and the tracked person walks naturally with interaction with the environment, including speed change and crowd crossing, and this scenario is also challenging for leg detection.
	
	\subsubsection{Evaluation Metrics}
	
	Following previous works \cite{2007Arras,2014Weinrich}, $TP$, $FP$, {\it precision rate}, and {\it recall rate} are employed as metrics to evaluate the leg detection performance.
	$TP$ is the number of leg clusters that are successfully detected.
	$FP$ corresponds to non-leg clusters classified to legs.
	The precision and recall rates are formulated as follows:
	\begin{equation}
		Precision = \frac{TP}{TP+FP}\quad Recall = \frac{TP}{TP+FN}
	\end{equation}
	where $FN$ corresponds to the leg clusters that fail to be detected.
	Furthermore, 
	to intuitively compare the performance of different leg detectors,
	the {\it precision-recall curve (PR curve)} is depicted in our experiment by taking $100$ values from $0.1$ to $0.9$ with equal intervals as the label assignment threshold $\beta$.
	
	\begin{table}[t]
		\centering
		\caption{COMPARISON WITH OTHER LEG DETECTOR}
		\label{tab:rf_methods}
		\begin{threeparttable}
			\begin{tabular}{p{2.2cm}<{\centering}p{1.0cm}<{\centering}p{0.6cm}<{\centering}p{0.6cm}<{\centering}p{1cm}<{\centering}p{1cm}<{\centering}}
				\toprule
				Method & Model & TP & FP & Precision & Recall \\
				\midrule
				Bellotto* {\it et al.}\cite{2009Bellotto} & - &14722 &18470 &55.88\% &44.27\% \\
				\specialrule{0em}{1pt}{1pt}
				Arras* {\it et al.}\cite{2007Arras} & AdaBoost &13474 &10080 &57.20\% &51.15\% \\
				\specialrule{0em}{1pt}{1pt}
				Leigh {\it et al.}\cite{2015Leigh} & SRF &21415 &5272 &80.25\% &81.29\%\\
				\specialrule{0em}{1pt}{1pt}
				Reis* {\it et al.}\cite{2019AJ} & PRF &22363 &3428 &86.71\% &84.89\%\\
				\specialrule{0em}{1pt}{1pt}
				Ours & MARF@3 &\bf{22955} &\bf{2825} &\bf{89.04\%} &\bf{87.14\%}\\
				\specialrule{0em}{1pt}{1pt}
				\bottomrule
			\end{tabular}
			\begin{tablenotes}
				\item Label assignment threshold $\beta$ is set to 0.5.
				\item * marks the re-implementation version.
			\end{tablenotes}
		\end{threeparttable}
	\end{table}
	
	To further evaluate the contribution of our MARF to people detection and tracking,
	we evaluate the performance on several indicators:
	the number of people successfully detected (Correct),
	the number of people missed (Miss),
	the number of wrong detected people (FPP),
	and widely-used Multi-Object Tracking Accuracy (MOTA) \cite{Bernardin2008Evaluating}.
	The MOTA is formulated as follows:
	\begin{equation}
		MOTA = 1 - \frac{\sum_i(ID_i + Miss_i + FPP_i)}{\sum_i {GT_i}}
	\end{equation}
	where $ID_i$ is the number of people whose ID changed from frame $i-1$ to frame $i$.
	$Miss_i$, $FPP_i$ are the Miss and the FPP in frame $i$.
	$GT_i$ is the number of people labeled in frame $i$.

	\subsubsection{Parameter Settings}
	$\epsilon$ in Eq.$($\ref{eq:gl}$)$ is a parameter to check global-local confidence conflict.
	To determine its value setting, we explore the effect of choices of parameter $\epsilon$, as shown in Fig. \ref{parameter}.
	The experiments are based on the best model configuration MARF@$3$, $\epsilon$ is set to $0.1$, and $\alpha$ is set to $0.6$.
	The default number of scale $K$ is set to $3$.

	\subsection{Comparison with Other Leg Detectors}
	\label{sec:rf_method}
	Table \ref{tab:rf_methods} shows the leg detection performances of several leg detectors.
	The detector \cite{2009Bellotto} is a heuristic method for leg recognition that directly outputs the cluster's label and we extend it for comparison of leg detection.
	The detector \cite{2007Arras} is an AdaBoost-based detector trained with the features they design.
	The detector \cite{2015Leigh} is a public SRF-based leg detector.
	And the detector \cite{2019AJ} is a PRF-based leg detector.
	It is first proposed to classify noisy astronomical \lwt{datasets}, and we re-implement it to classify legs for comparison.
	Our leg detector is the proposed MARF@$K$ model, and the $K$ is set to 3.
	
	\begin{figure}[t]
		\begin{center}
			\includegraphics[width=1\linewidth]{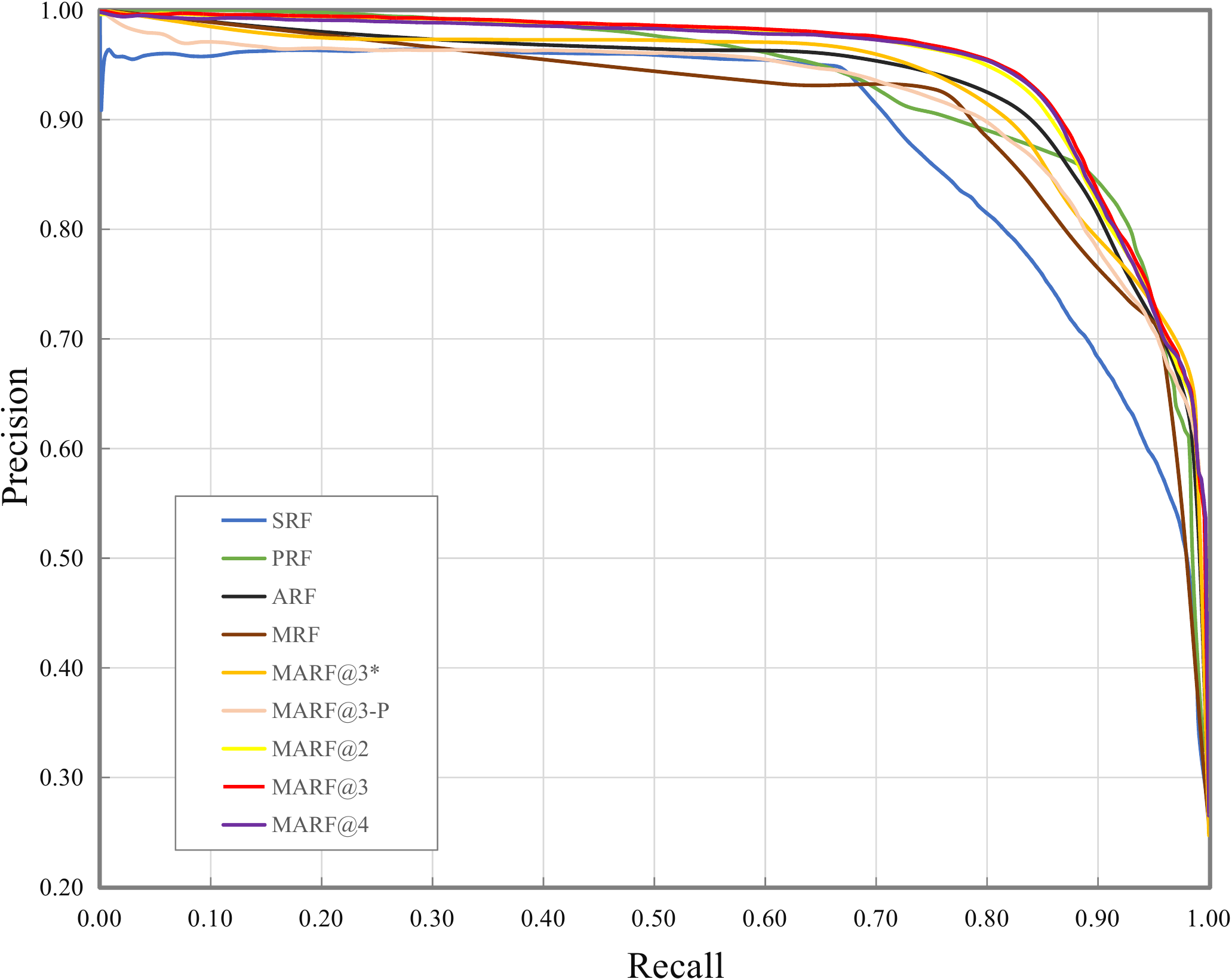}
			\caption{Precision-Recall curves of different models on {\it Moving Legs} dataset.}
			\label{pr_curve}
		\end{center}
	\end{figure}
	\begin{table}[t]
		\centering
		\caption{ABLATION STUDY ON MOVING LEGS}
		\label{ablation}
		\begin{threeparttable}
			\begin{tabular}{p{2.0cm}<{\centering}p{1.1cm}<{\centering}p{1.1cm}<{\centering}p{1.1cm}<{\centering}p{1.1cm}<{\centering}}
				\toprule
				Method & TP &FP & Precision & Recall \\
				\midrule
				MRF       &22674                   &4956                       &82.06\%                    &86.07\%     \\
				\specialrule{0em}{1pt}{1pt}
				ARF       &22697                   &3782     &85.72\%  &86.16\%     \\
				\specialrule{0em}{1pt}{1pt}
				MARF@3      &\bf{22955}  &\bf{2825}      &\bf{89.04\%}   &\bf{87.14\%}  \\
				\specialrule{0em}{1pt}{1pt}
				\bottomrule
			\end{tabular}
			\begin{tablenotes}
				\item Label assignment threshold $\beta$ is set to 0.5.
			\end{tablenotes}
		\end{threeparttable}
	\end{table}
	
	Intuitively,
	our method achieves 8.79\% precision improvement and 5.85\% recall improvement over the SRF-based detector.
	Furthermore, the FP of ours is only 53.58\% of that of the SRF-based method.
	There is no doubt that our method has a considerable boost in performance compared with the SRF-based method,
	as the proposed method has immunity to noise that is inevitable in 2D Laser.
	Moreover,
	although the PRF-based detector also has specific immunity to noise,
	ours achieves 2.33\% precision improvement, 2.25\% recall improvement, 17.6\% FP decrease over the PRF-based detector.
	The reason is that the proposed method avoids terminating training too early and fully utilizes the multi-scale characteristics of the leg cluster.
	For TP, our method also achieves the best performance.
	Fig. \ref{pr_curve} compares their performances more comprehensively.
	Intuitively, our method outperforms other methods at almost all thresholds.
	Although PRF achieves a weak advantage when recall ranges from 0.9 to 0.95,
	its precision has a more apparent decline when recall ranges from 0.6 to 1.
	In contrast,
	ours achieves a great advantage when recall ranges from 0.6 to 0.9 while maintaining a high and stable precision,
	which further proves the effectiveness of our method.

	\subsection{Ablation Study}
	\label{sec:ablation}
	To demonstrate the significance of the adaptive-switch decision tree and multi-scale structure respectively,
	we disable each proposed module to evaluate our contributions.
	Note that all variants are derived from MARF@3 and evaluated at {\it Moving Legs} dataset.
	
	\subsubsection{Contribution of Adaptive-switch Decision Tree}
	
	The multi-scale random forest (MRF) is the standard random forest with \lwt{a} multi-scale structure, replacing the adaptive-switch decision tree with \lwt{a} standard decision tree.
	Table \ref{ablation} shows the variants of the proposed leg detector.
	MARF@3 achieves 6.98\% precision improvement, 1.07\% recall improvement, 43\% FP decrease over MRF,
	which is due to the effectiveness of the proposed adaptive-switch decision tree.
	It is noteworthy that the significant drop in false-positive detection helps the follow-up work to eliminate non-leg clusters.
	The PR curves, shown in Fig. \ref{pr_curve}, also illustrates the necessity of adaptive-switch decision tree. 
	The reason is that the proposed method measures the global-local confidence conflict and employs dichotomous nodes to make a more reasonable split,
	promoting the discriminative capability of each classifier in the forest.
	
	\subsubsection{Contribution of Multi-scale Structure}
	
	Discarding the multi-scale structure, ARF is a single-scale random forest composed of adaptive-switch decision trees.
	By comparing MARF@3 and ARF in Table \ref{ablation},
	it can be seen that MARF@3 achieves 3.32\% precision improvement, 0.98\% recall improvement, 25.3\% FP decrease over MRF.
	Moreover, the PR curve shown in Fig. \ref{pr_curve} depicts a similar observation.
	The intrinsic reason is that \lwt{the} multi-scale structure \lwt{contributes} to the robustness of the exploitation of multi-scale leg features,
	and overlapping fusion boosts leg detection accuracy in all scales.

	\begin{table}[t]
		\centering
		\caption{COMPARISON OF MULTI-SCALE STRUCTURE}
		\label{analysis}
		\begin{threeparttable}
			\begin{tabular}{p{2.0cm}<{\centering}p{1cm}<{\centering}p{1cm}<{\centering}p{1.2cm}<{\centering}p{1.2cm}<{\centering}}
				\toprule
				Method & TP &FP & Precision & Recall \\
				\midrule
				MARF@2 &22932  &2893 &88.80\% &87.08\%  \\
				\specialrule{0em}{1pt}{1pt}
				MARF@3 &\bf{22955}  &\bf{2825} &\bf{89.04\%} &\bf{87.14\%}  \\
				\specialrule{0em}{1pt}{1pt}
				MARF@4 &22943  &2917 &88.72\% &87.09\% \\
				\specialrule{0em}{1pt}{1pt}
				MARF@3$^*$  &22843 &4520 &83.48\% &86.71\% \\
				\specialrule{0em}{1pt}{1pt}
				MARF@3-P &22427 &3663 &85.96\% &85.13\%  \\
				\bottomrule
			\end{tabular}
			\begin{tablenotes}
				\item Label assignment threshold $\beta$ is set to 0.5.
			\end{tablenotes}
		\end{threeparttable}
	\end{table}
	\subsection{Model Details Analysis}
	\label{details}
	Previous comparisons illustrate the performance and advantage of our method.
	However, there are still some details of module design that are worthy of being further discussed.
	In this section, we first discuss how the setting of scale number $K$ influences the performance of our model.
	Subsequently, another distinct multi-scale structure is employed to compare with the proposed structure.
	At last, to illustrate the contribution of the proposed dichotomous node,
	based on the best multi-scale structure, we replace the dichotomous node with the probabilistic node for comparison.
	
	\subsubsection{Analysis of the Scale Number}
	In this part, three multi-scale structures with different scale numbers are designed for experimental comparison.
	We refer \lwt{to} the MARF with $K$ scales forests as MARF@$K$,
	and the scale details are presented as follows:
	
	\begin{itemize}
		\item MARF@2: [0m, 6m), [6m, +$\infty$)
		\item MARF@3: [0m, 3m), [3m, 6m), [6m, +$\infty$)
		\item MARF@4: [0m, 1.5m), [1.5m, 3m), [3m, 6m), [6m, +$\infty$)
	\end{itemize}
	In our method,
	three ARFs in MARF@3 are employed to predict the cluster in three ranges:
	[0m, +$\infty$), [3m, +$\infty$), and [6m, +$\infty$).
	The results are shown in Table \ref{analysis}, MARF@3 outperforms MARF@2 and MARF@4 and obtains the best performance. 
	The results adequately illustrate that the depth discretization of MARF@3 is more suitable for leg detection in different distances.
	
	\begin{table}[t]
		\centering
		\caption{COMPARISON OF LEAF NODE}
		\label{attribute}
		\begin{threeparttable}
			\begin{tabular}{p{1.5cm}<{\centering}p{2cm}<{\centering}p{3.5cm}<{\centering}}
				\toprule
				Method & Average Depth & Average Leaf Node Number  \\
				\midrule
				MARF@3  &17.3            &264.6 \\
				\specialrule{0em}{1pt}{1pt}
				MARF@3-P    &13.8  &196.7  \\
				\bottomrule
			\end{tabular}
		\end{threeparttable}
	\end{table}

	\begin{table}[t]
		\caption{RESULTS ON PEOPLE TRACKING DATASET}
		\label{result_tb}
		\begin{center}
			\begin{threeparttable}
				\begin{tabular}{cccccc}
					\toprule
					Dataset & Tracker & Correct  & Miss & FPP & MOTA \\
					\midrule
					\multirow{3}{*}{SCENARIO \uppercase\expandafter{\romannumeral1}} & leg\_detector\cite{Lu2013Towards} &478 & 1605 & 28 & 19.2\% \\
					\specialrule{0em}{1pt}{1pt}
					~ & Joint Leg \cite{2015Leigh} &711  & 1372 & 9 & 33.2\%\\
					\specialrule{0em}{1pt}{1pt}
					~ & MARF-based & \bf{743} & \bf1341 & 9 & \bf35.05\% \\
					\midrule
					\multirow{3}{*}{SCENARIO \uppercase\expandafter{\romannumeral2}} & leg\_detector\cite{Lu2013Towards} &164  & 481 & 294 & -22.5\% \\
					\specialrule{0em}{1pt}{1pt}
					& Joint Leg \cite{2015Leigh} &165  & 480 & 97 & 10.2\% \\
					\specialrule{0em}{1pt}{1pt}
					& MARF-based & \bf201 & \bf 444 & \bf51 & \bf23.01\%\\
					\midrule
					\multirow{3}{*}{SCENARIO \uppercase\expandafter{\romannumeral3}} & leg\_detector \cite{Lu2013Towards} &15621  & 3425 & - & -\\
					\specialrule{0em}{1pt}{1pt}
					& Joint Leg \cite{2015Leigh} &18561  & 485 &- & -\\
					\specialrule{0em}{1pt}{1pt}
					& MARF-based &\bf18894  &\bf152 & - & -\\
					\bottomrule
				\end{tabular}
				\begin{tablenotes}
					\footnotesize
					\item[-] Due to only part of people are labeled on the ground truth in DATASET \uppercase\expandafter{\romannumeral3} except tracked people, FP and MOTA are ignored \cite{2015Leigh}.
				\end{tablenotes}
			\end{threeparttable}
		\end{center}
	\end{table}

	\begin{figure*}[t]
		\begin{center}
			\includegraphics[width=0.97\linewidth]{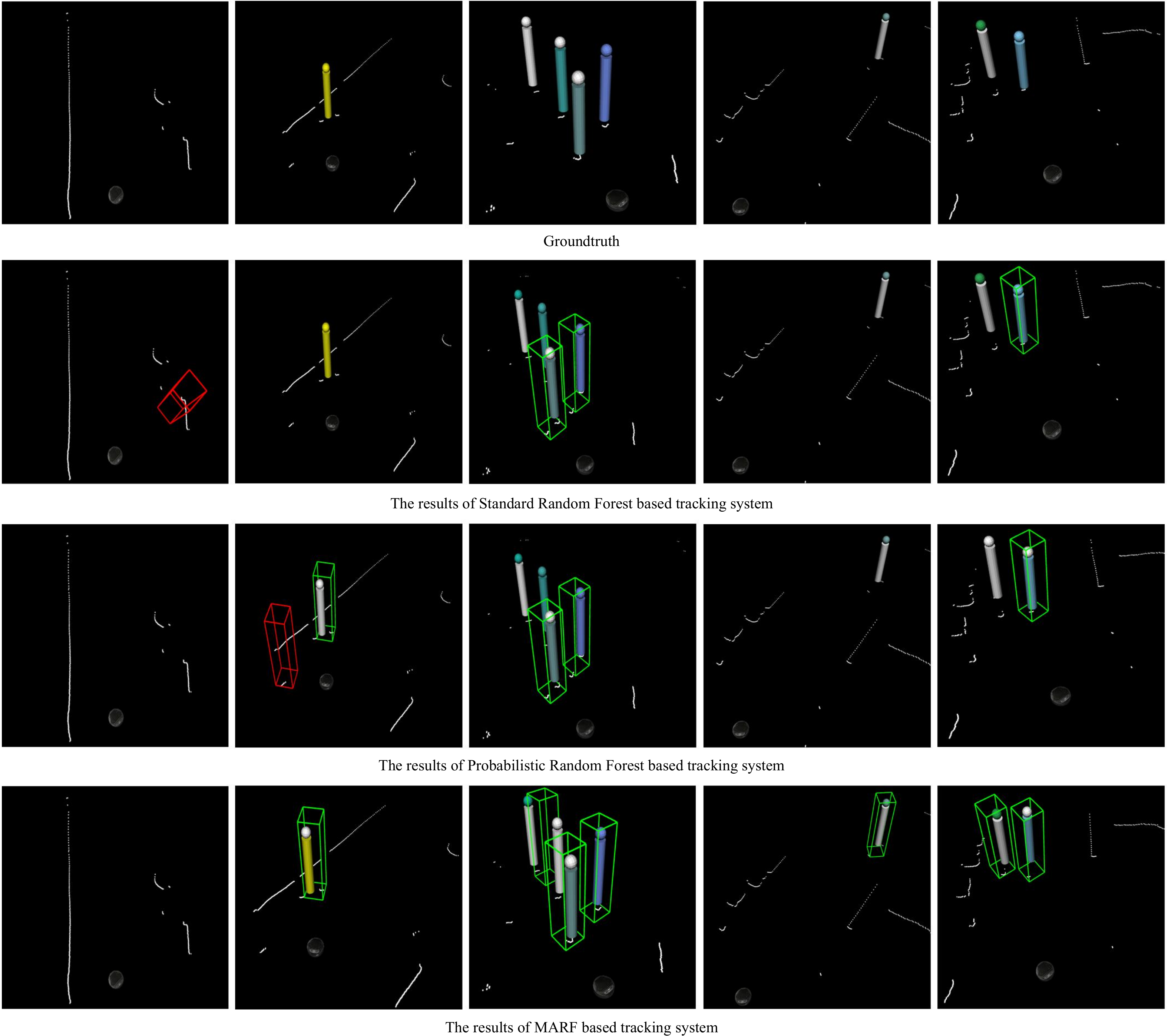}
			\caption{Qualitative results of the evaluated methods for people detection. Five scenes in the first column represent different point cloud frames.
				The following columns show the people detection results presented by the SRF-based method \cite{2015Leigh}, the PRF-based method \cite{2019AJ} and our method (MARF@3), respectively.
				The cylinder represents ground truth people,
				red boxes denote the false positive detection,
				and green boxes denote the true positive detection.}
			\label{fig:qualitative}
		\end{center}
	\end{figure*}
	
	\subsubsection{Analysis of the Multi-scale Structure}
	To illustrate the rationality of the proposed overlapping multi-scale structure,
	we construct a non-overlapping structure for comparison,
	namely, MARF@3$^*$.
	In MARF@3$^*$,
	there are no intersection scales between each ARFs,
	e.g., three ARF focus on the clusters ranges $(0$m-$3$m$]$, $(3$m-$6$m$]$, and $(6$m-$\infty)$, respectively.
	Each scale forest is trained with clusters in the corresponding scales and predicts the legs independently.
	The results are shown in Table \ref{analysis}.
	MARF@3$^*$ has a large decline in performance compared with MARF@3,
	which proves the advantage of overlapping \lwt{structures} in the leg detection task.
	The main reason is that the overlapping structure shares training clusters between scales.
	This enriches the scale variety of training \lwt{sets} and contributes to the generalization of the decision tree, and even the whole model.

	\subsubsection{Analysis of the Dichotomous Node}
	We build a variant of the adaptive-switch decision tree to illustrate the effectiveness of the proposed dichotomous node.
	In this variant,
	called MARF@3-P,
	we employ the probabilistic node proposed in \cite{2019AJ} to replace the dichotomous node in MARF@3.
	The comparison is shown in Table \ref{analysis}.
	MARF@3 achieves 3.08\% precision improvement, 3.01\% recall improvement, and 22.9\% FP decrease over MARF@3-P.
	The main reason is that clusters passed by the probabilistic nodes are difficult to go deeper,
	which generates more shallow-layer leaf nodes and leads to insufficient training.
	The average depth of leaf nodes is shown in Table \ref{attribute},
	which proves our inference.
	
	\subsection{Further Application on People Detection and Tracking}
	\label{further}
	Due to the dominant effect of \lwt{the} leg detector in people detection and tracking,
	we further evaluate the proposed detector's contribution to this task.
	In more detail,
	we apply MARF@3 to the Joint Leg \cite{2015Leigh} people tracking framework.
	The Joint Leg employs SRF to detect legs from 2D laser scans at first.
	Then they match legs as leg-pairs and apply a heuristic method to detect people. 
	Finally, the method uses a Kalman Filter to track people in consecutive laser scans.
	For reasonable evaluation and comparison,
	MARF@3 is substituted for SRF in Joint Leg to detect legs.

	\subsubsection{Quantitative Results}
	As shown in Table \ref{result_tb}, our
	MARF-based system achieves the best performance in Correct, Miss,  and FPP for people detection.
	In SCENARIO \uppercase\expandafter{\romannumeral1},
	since MARF detects more long-distance people and achieves a noticeable improvement in correct people detection,
	ours achieves 4.3\% Correct improvement and 2.3\% Miss decrease over Joint Leg \cite{2015Leigh}.
	In SCENARIO \uppercase\expandafter{\romannumeral2},
	contrasting to SRF generating more FP,
	our approach achieves 17.9\% Correct improvement, 7.5\% Miss decrease, and 47.4\% FPP decrease over the Joint Leg \cite{2015Leigh}.
	In SCENARIO \uppercase\expandafter{\romannumeral3},
	our MARF-based system achieves 1.8\% Correct improvement and 68.7\% Miss decrease over Joint Leg \cite{2015Leigh}.
	These results further prove that the robust classification capability of MARF conduces to detect people.
	
	For people tracking,
	benefiting from leg detection in consecutive laser data frames,
	robust people detection contributes to a higher MOTA score.
	MARF achieves 1.85\% MOTA improvement in SCENARIO \uppercase\expandafter{\romannumeral1} and 12.81\% MOTA improvement in SCENARIO \uppercase\expandafter{\romannumeral2}.
	Therefore the results can further prove the advance of MARF.

	\subsubsection{Qualitative Results}
	Fig. \ref{fig:qualitative} shows the qualitative result tested on the {\it People Tracking} dataset.
	The first row shows the scenario that only non-leg clusters occur nearby the laser scanner.
	The SRF-based method \cite{2015Leigh} mistakenly recognizes non-leg clusters as the people, while our MARF-based system and PRF-based system correctly predict it.
	The second row shows a person standing nearby the wall, and an ambiguous cluster is similar to the leg.
	The SRF-based method can hardly distinguish both leg and non-leg clusters.
	The PRF-based method correctly recognizes the people, however, it also induces the false positive detections. 
	Similarly, our MARF-based system can effectively avoid fake leg clusters and accurately determine people in the fifth scenario.
	The third row depicts the scenario with crowded people. 
	Ours detects the most people,
	but there are still some challenges in such occluding conditions.
	The fourth row shows a long-distance people which the SRF-based and the PRF-based approaches can hardly detect.
	While our MARF-based system successfully detects this long-distance people.
	The fifth row shows two persons that are far away from the robot.
	Although the two persons are near each other,
	our method successfully detects them all.
	
	\begin{table}[t]
		\centering
		\caption{TIME COMPARISON WITH OTHER RANDOM FORESTS}
		\label{tab:rf_time}
		\begin{threeparttable}
			\begin{tabular}{p{1.0cm}<{\centering}p{1cm}<{\centering}p{1cm}<{\centering}p{0.8cm}<{\centering}p{1.2cm}<{\centering}p{1.3cm}<{\centering}}
				\toprule
				Model & Precision & Recall & Code & s /image & Used Node \\
				\midrule
				\specialrule{0em}{1pt}{1pt}
				SRF & 80.25\% & 81.29\% & OpenCV & 0.0079 & - \\
				\specialrule{0em}{1pt}{1pt}
				PRF & 86.71\% & 84.89\% & C++ & 0.0088 & 526 \\
				\specialrule{0em}{1pt}{1pt}
				MARF@3 & 89.04\% & 87.14\% & C++ & 0.0138 & 4091 \\
				\specialrule{0em}{1pt}{1pt}
				\bottomrule
			\end{tabular}
			\begin{tablenotes}
				\item Label assignment threshold $\beta$ is set to 0.5.
			\end{tablenotes}
		\end{threeparttable}
	\end{table}

	\subsubsection{Efficiency and Speed Comparison}
	The efficiency of algorithm is also concerned,
	especially for the application on low-cost robot \lwt{platforms}.
	We record time cost on a ThinkPad E550 laptop with CPU i5-5500U, Ubuntu 16.04LTS OS and 8GB RAM.
	Owing to the dichotomous node,
	the prediction of each cluster is outputted by more leaf nodes in MARF@3 and is expected to cost more time.
	However, during leg detection, MARF@3 can still process laser scans at almost 62Hz, satisfying the real-time detection requirement.
	Moreover, we also argue that an optimization of the implementation of approach can further improve the running efficiency.
	While applying to the people detection and tracking framework \cite{2015Leigh},
	the MARF-based system achieves the same efficiency \lwt{as} the SRF-based approach, with \lwt{a} speed 7.2Hz.
	It proves that data association for people detection and tracking is the computational bottleneck in current work.
	That is also the next work we aim to study and optimize in the future.

	To clarify our method more comprehensively,
	we statistic the time cost of the proposed random forest and the existing random forest models,
	which is demonstrated in Table \ref{tab:rf_time}.
	It can be seen that the speed of MARF is relatively slower among all models.
	The reason is that,
	in PRF, each point cluster passes through 526 nodes on average,
	while 4091 nodes for MARF.
	This phenomenon is consistent with the Fig. 2, namely,
	MARF is usually deeper than PRF,
	so there are more nodes of MARF involved in the calculation.
	In addition,
	SRF is implemented with OpenCV API,
	thus we cannot fairly compare MARF to SRF.
	In the future, 
	how to obtain better performance with the same number of nodes is also a question worth studying.
	
	\begin{figure}[t]
		\begin{center}
			\includegraphics[width=1\linewidth]{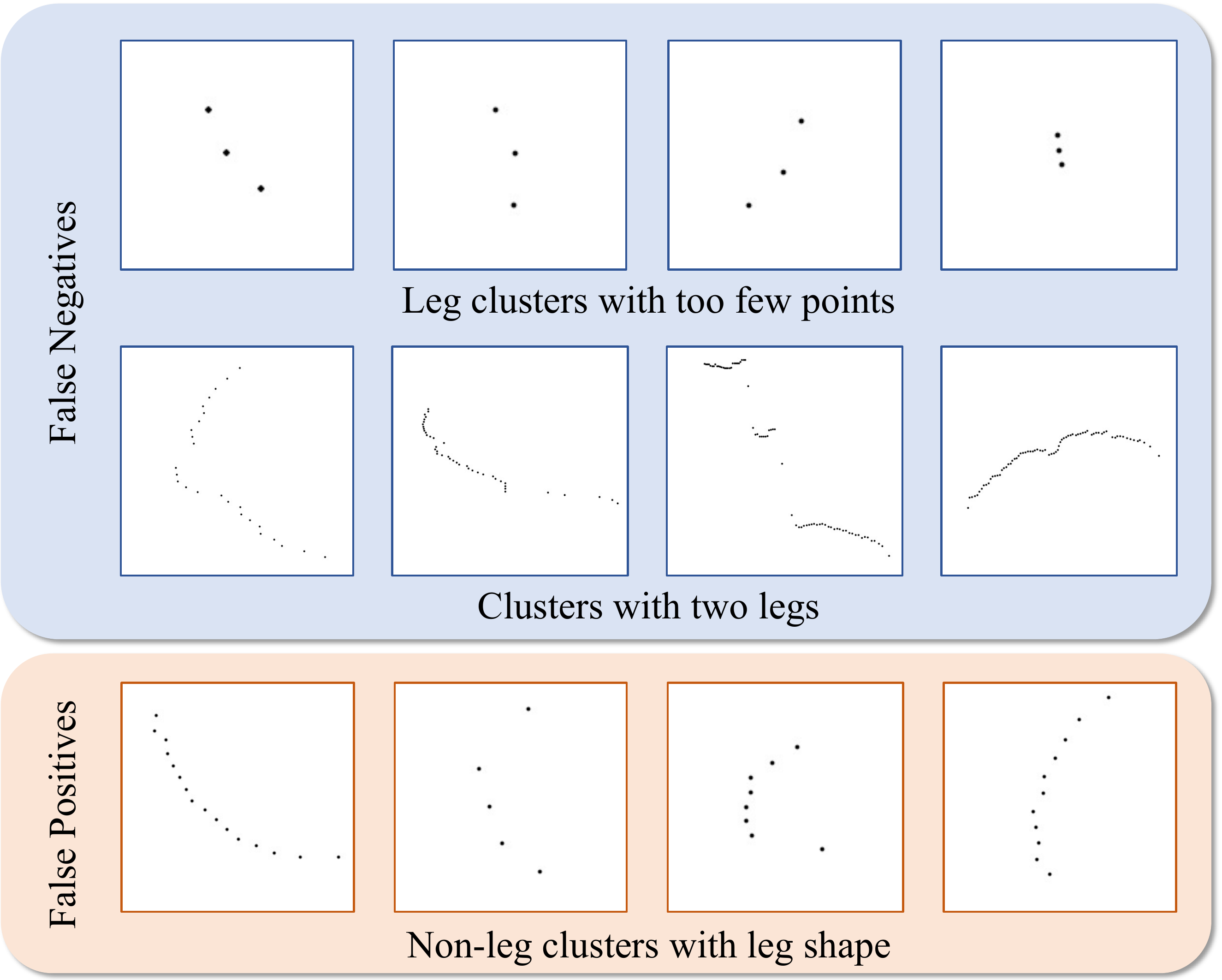}
			\caption{Failure cases of our method.}
			\label{fig:failure}
		\end{center}
	\end{figure}
	
	\subsection{Failure Cases}
	Fig. \ref{fig:failure} illustrates several failure cases of our method.
	Firstly, our method still cannot completely discover the legs with too few laser points. This issue is also a common difficulty \lwt{for} other leg detectors.
	Secondly, since the proposed method follows a single-leg detection pipeline that first segments the whole 2D point cloud and then classifies each point \lwt{cluster},
	it is inevitable for our model to classify some wrong-segmented clusters,
	such as a cluster indicating leg pair.
	These leg-pair clusters are so different from \lwt{the} single leg,
	thus they usually obtain very low confidence, as shown in the second row of Fig. \ref{fig:failure}.
	Thirdly, another unavoidable issue is the false positive detection of the static non-leg point cluster with leg shape in the scene.
	As shown in the third row of Fig. \ref{fig:failure},
	these clusters belong to static objects at different distances.
	In general, these failure cases are basically due to the lack of information in the 2D laser point cloud.
	In the future, we are going to introduce richer information to address this issue,
	e.g. combining multiple laser scans or incorporating consecutive laser scans to extract dynamic leg \lwt{clusters}.

	\section{CONCLUSIONS}
	\label{sec:conclusions}
	This paper focuses on a new problem,
	i.e., the conflict between global and local confidence of cluster features in 2D laser-based leg detection,
	and proposes an approach to alleviate this problem.
	Moreover, a Multi-scale Adaptive-switch Random Forest is proposed to enhance the classification ability of multi-scale leg clusters.
	We carefully evaluate our approach on the dataset and further apply it to people detection and tracking.
	Expanding experiments and comparative analysis validate the effectiveness of the proposed method.
	
	\begin{figure}[t]
		\centering
		\includegraphics[width=1\linewidth]{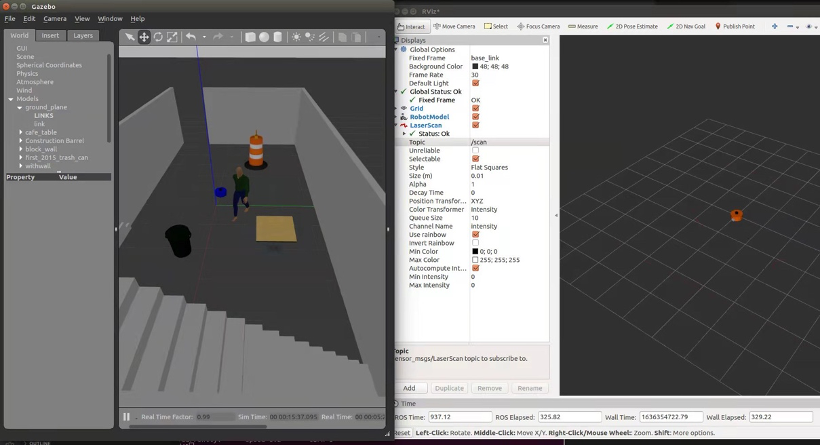}
		\caption{The experimental environment for measuring feature noise.}
		\label{fig:envr}
		\vspace{-3mm}
	\end{figure}
	
	\begin{figure}[t]
		\centering
		\includegraphics[width=0.95\linewidth]{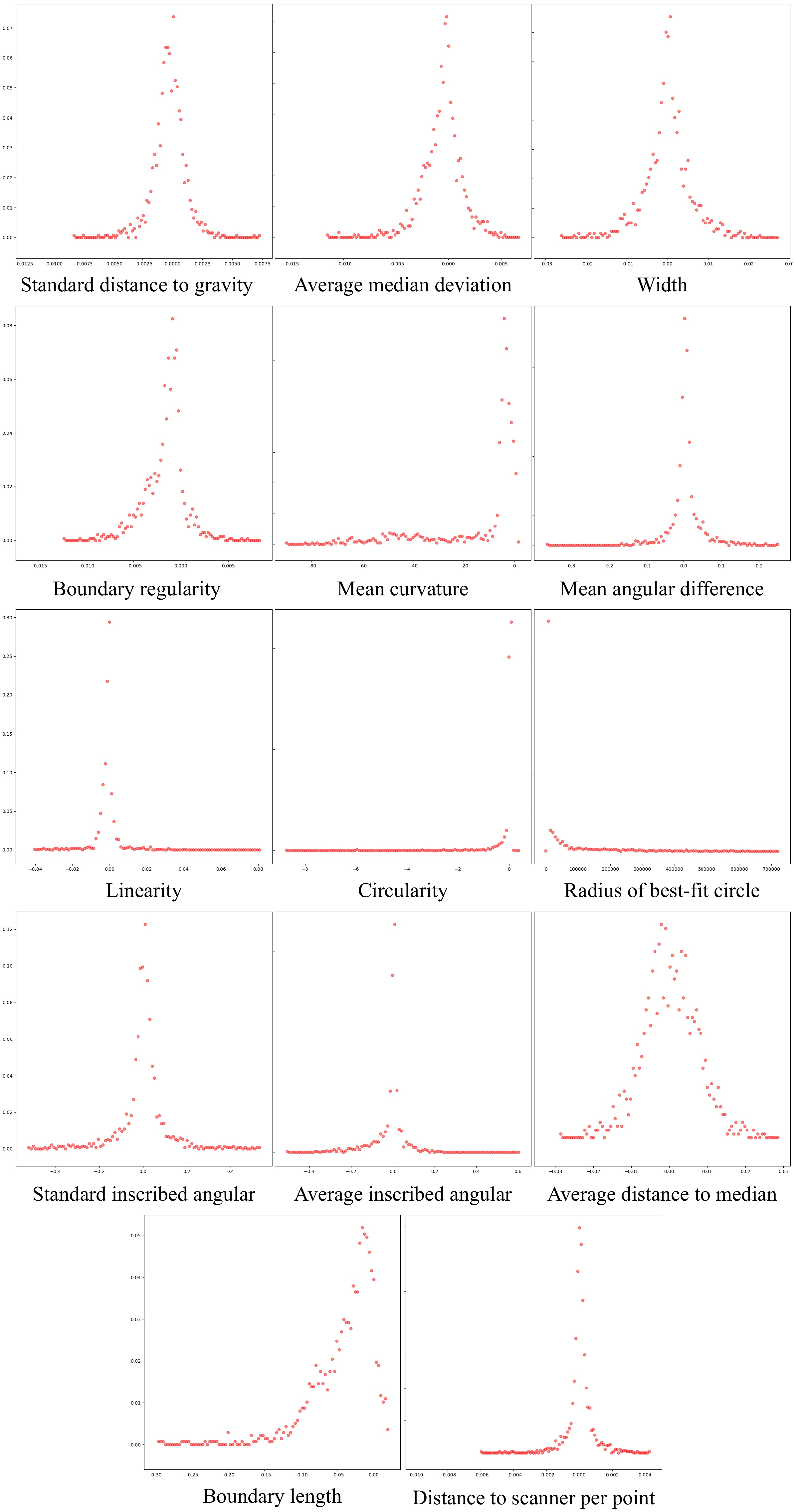}
		\caption{The noise visualization of the features based on Gaussian noisy laser scan.}
		\label{fig:featnoise1}
	\end{figure}
	
	\begin{figure} [t]
		\centering
		\includegraphics[width=0.95\linewidth]{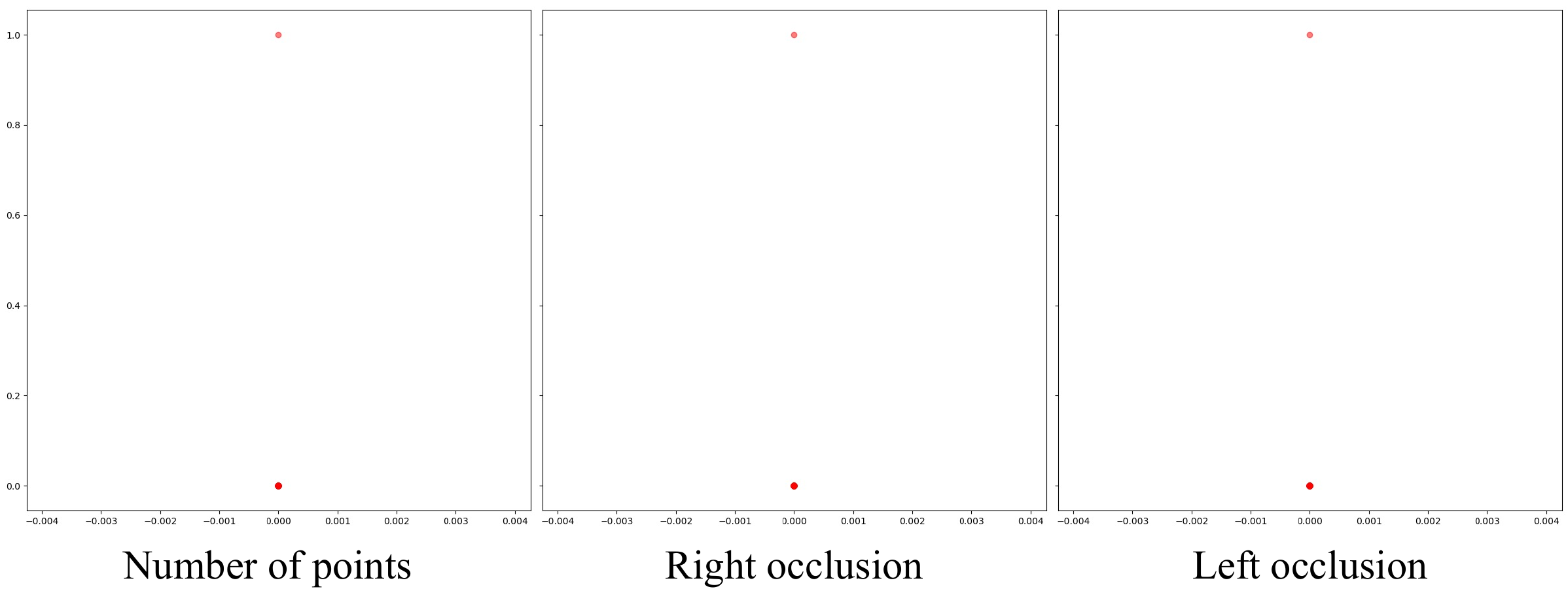}
		\caption{Noise visualization of features not affected by noise.}
		\label{fig:featnoise2}
	\end{figure}

	\appendix[Feature Noise Distribution of 2D Laser Scan]
	In this paper,
	we make a hypothesis that most of the employed features suffer from Gaussian noise.
	Next, we give the specific proof.

	Theoretically,
	the measurement error of 2D laser scan satisfies the distribution of white Gaussian noise \cite{2007Development,2006Geometrical}.
	In our method,
	the feature calculation encodes this Gaussian noise and laser point cloud together.
	Naturally,
	if the point cloud suffers from great noise,
	the deviation between the obtained feature and the true value would be large, and vice versa.
	Therefore,
	we make the hypothesis that the noise contained in each feature satisfies the Gaussian distribution.

	To further prove the reasonability of our hypothesis,
	we have conducted an experiment to evaluate the noise of each employed feature.
	Firstly,
	we build a room without any noise by using the Gazebo simulator,
	with a moving person, several obstacles, and a mobile robot mounted a 2D laser scanner,
	as shown in Fig. \ref{fig:envr}.
	Secondly,
	we record a large number of noise-free 2D point clouds by this laser scanner,
	and meanwhile, control the robot to move in this room.
	Finally, we add white Gaussian noise to the point cloud and compare it to the original point cloud in the feature domain.
	The errors of each feature are given in Fig. \ref{fig:featnoise1} and Fig. \ref{fig:featnoise2}.

	Intuitively,
	except for three features that are not related to noise,
	other features are mostly affected.
	The errors satisfy the Gaussian distribution.
	Thus, we believe that this hypothesis is reasonable.

	\ifCLASSOPTIONcaptionsoff
	\newpage
	\fi
	
	\balance
	\bibliographystyle{IEEEtran}
	\bibliography{reference}
	
\end{document}